# Fingerprint Gender Classification using Wavelet Transform and Singular Value Decomposition

**Gnanasivam P[1], and Dr. Muttan S[2]**

**[1] Department of Electronics and Communication Engineering, Agni College of Technology**
**Chennai, Tamilnadu-603103, India**
***pgnanasivam@yahoo.com***

**[2] Center for Medical Electronics, Department of ECE, College of Engineering, Anna University**
**Chennai, Tamilnadu-600025, India**
**Muthan_s@annauniv.edu**

## Abstract

A novel method of gender Classification from fingerprint is proposed based on discrete wavelet transform (DWT) and singular value decomposition (SVD). The classification is achieved by extracting the energy computed from all the sub-bands of DWT combined with the spatial features of non-zero singular values obtained from the SVD of fingerprint images. K nearest neighbor (KNN) used as a classifier. This method is experimented with the internal database of 3570 fingerprints finger prints in which 1980 were male fingerprints and 1590 were female fingerprints. Finger-wise gender classification is achieved which is 94.32% for the left hand little fingers of female persons and 95.46% for the left hand index finger of male persons. Gender classification for any finger of male persons tested is attained as 91.67% and 84.69% for female persons respectively. Overall classification rate is 88.28% has been achieved.

## 1. Introduction

Gender and Age information is important to provide investigative leads for finding unknown persons. Existing methods for gender classification have limited use for crime scene investigation because they depend on the availability of teeth, bones, or other identifiable body parts having physical features that allow gender and age estimation by conventional methods. Various methodologies has been used to identify the gender using different biometrics traits such as face, gait, iris, hand shape, speech and fingerprint. In this work, gender and age of a person is identified from the fingerprint using DWT and SVD. The science of fingerprint has been used generally for the identification or verification of person and for official documentation. Fingerprint analysis plays a role in convicting the person responsible for an audacious crime. Fingerprint has been used as a biometric for the gender and age identification because of its unique nature and do not change throughout the life of an individual [1].

In fingerprint, the primary dermal ridges (ridge counts) are formed during the gestational weeks 12-19 and the resulting fingerprint ridge configuration (fingerprint) is fixed permanently [2-3].Ridges and their patterns exhibit number of properties that reflect the biology of individuals. Fingerprints are static and its size and shape changes may vary with age but basic pattern of the fingerprint remains unchanged. Also, the variability of epidermal ridge breadth in humans is substantial [4]. Dermatoglyphic features statistically differ between the sexes, ethnic groups and age categories [5]. It is proved by various researchers; a fingerprint can be processed for the sex determination [6-11]. Thus the variability in sex and age with size, shape and the ridge width of fingerprints helps for the study. Gender and age determination of unknown can guide investigators to the correct identity among the large number of possible matches. Figure 1 illustrates the process of DWT and SVD based gender and classification system.

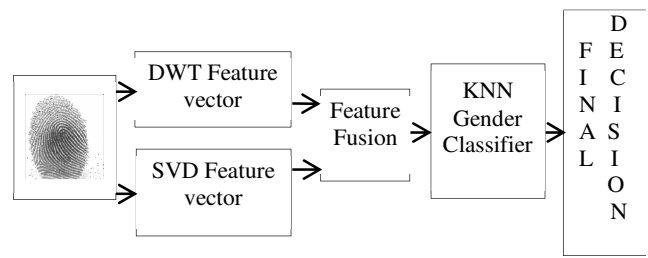

Fig. 1 DWT and SVD based gender classification system.

According to 'Police-reported crime statistics in Canada, 2010'[12], Crimes tend to be disproportionately committed by youth and young adults and the rate of those accused of a *Criminal Code* offence peaked at 18 years of age and generally decreased with increasing age. According to

'crime in India statistics-2010', published by National Crime Records Bureau [13], the crime rate is higher for the age range of 18 to 44 and decreases after 44. The crime done by female is only 4-8% in comparison with the crime done by male.

Wavelet transform is a popular tool in image processing and computer vision because of its complete theoretical framework, the great flexibility for choosing bases and the low computational complexity [14]. As wavelet features has been popularized by the research community for wide range of applications including fingerprint recognition, face recognition and gender identification using face, authors have confirmed the efficiency of the DWT approach for the gender identification using fingerprint.

The SVD approach is selected for the gender discrimination because of its good information packing characteristics and potential strengths in demonstrating results. The SVD method is considered as an information-oriented technique since it uses principal components analysis procedures (PCA), a form of factor analysis, to concentrate information before examining the primary analytic issues of interest [15]. K-nearest neighbors (KNN), gives very strong consistent results. It uses the database which was generated in the learning stage of the proposed system and it classifies genders of the fingerprints.

The outline of this paper is as follows: we review the previous approaches for sex determination using fingerprint in section 2, followed by discussions of fingerprint feature extraction in Section 3; we then proposed the gender classification using fingerprint features in Section 4; the experimental results are presented in Section 5; Section 6 comes to the conclusion and future work.

## 2. Previous Approaches

Gender and age classification can be me made using the spatial parameters or frequency domain parameters or using the combination of both. Most of the findings are based on the spatial domain analysis and few were based on the frequency domain. Earlier work on gender classification based on the ridge density shows that the ridge density is greater for female than male [7,8, 16,17] and [9] analyzed fingerprints of bagathas a tribal population of Andhra Pradesh (India) and showed the evident that the males showing higher mean ridge counts than females. Importance of ridge distance [18, 19] and ridge period [20] and ridge frequency [21] measurements as spatial parameters in the context of fingerprint gender classification are explained. Except few papers, fingerprint gender identification is made by manual measurements from the inked fingerprints. Many studies were carried out for the human face gender classification by using frequency domain and various classifiers [14, 22-29]. Only few efforts have been made for the gender classification through fingerprint.

## 3. Fingerprint feature extraction

Feature extraction is a fundamental pre-processing step for pattern recognition and machine learning problems. In the proposed method, the energy of all DWT sub-bands and non-zero singular values obtained from the SVD of fingerprint image are used as features for the classification of gender. In this section, DWT and SVD based fingerprint feature extractions are described.

### 3.1 DWT Based Fingerprint Feature extraction

Wavelets have been used frequently in image processing and used for feature extraction, de-noising, compression, face recognition, and image super-resolution. Two dimensional DWT decomposes an image into sub-bands that are localized in frequency and orientation. The decomposition of images into different frequency ranges permits the isolation of the frequency components introduced by "intrinsic deformations" or "extrinsic factors" into certain sub-bands. This process results in isolating small changes in an image mainly in high frequency sub-band images. Hence, DWT is a suitable tool to be used for designing a classification system.

The 2-D wavelet decomposition of an image is results in four decomposed sub-band images referred to as low–low (LL), low–high (LH), high–low (HL), and high–high (HH). Each of these subbands represents different image properties. Typically, most of the energy in images is in the low frequencies and hence decomposition is generally repeated on the LL sub band only (dyadic decomposition). For k level DWT, there are (3*k) + 1 sub-bands available. The energy of all the sub-band coefficients is used as feature vectors individually which is called as sub-band energy vector (E). The energy of each sub-band is calculated by using the equation (1).

$$E_k = \frac{1}{RC} \sum_{i=1}^{R} \sum_{j=1}^{C} |x_k(i,j)| \qquad (1)$$

Where $x_k(i,j)$ is the pixel value of the $k$th sub-band and R, C is width and height of the sub-band respectively.

Figure 2 shows the block diagram of the frequency feature extraction by using DWT. The input fingerprint image is first cropped and then decomposed by using the DWT. For level 1, number of subbands are 4 and 3 subbands are added for each next levels. Thus the increase in levels of DWT increases the features.

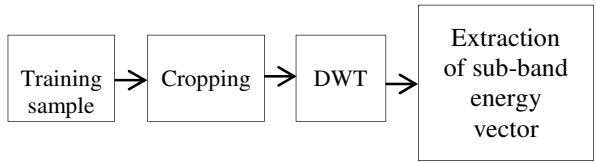

Fig. 2 DWT based fingerprint feature extraction

## 3.2 SVD based Fingerprint Feature extraction

The Singular Value Decomposition (SVD) is an algebraic technique for factoring any rectangular matrix into the product of three other matrices. Mathematically and historically, it is closely related to Principal Components Analysis (PCA). In addition it provides insight into the geometric interpretation of PCA. As noted previously, the SVD has long been considered fundamental to the understanding of PCA.

The SVD is the factorization of any $k \ X \ p$ matrix into three matrices, each of which has important properties. That is, any rectangular matrix A of k rows by p columns can be factored into U, S and V by using the equation (2).

$$A = U \ S \ V^T \qquad (2)$$

Where

$$U = AA^T \qquad (3)$$

$$V = A^T A \qquad (4)$$

And S is a k X p diagonal matrix with r non-zero singular values on the diagonal, where r is the rank of A. Each singular value is the square root of one of the Eigen values of both $AA^T$ and $A^T A$. The singular values are ordered so that the largest singular values are at the top left and the smallest singular values are at the bottom right, i.e., $s_{1,1} \geq s_{2,2} \geq s_{3,3}$ etc.

Among the three rectangular matrices, S is a diagonal matrix which contains the square root Eigen values from U or V in descending order. These values are stored in a vector called Eigen vector (V). As the internal database contains images of size 260x300 pixels, the feature vector of SVD is of the size 1x260. The spatial feature extraction by using SVD is shown in Figure 3.

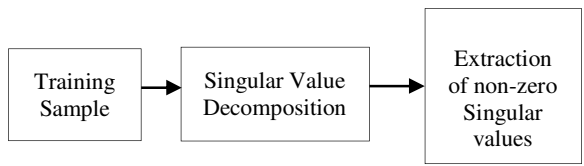

Fig. 3 SVD based fingerprint feature extraction

## 4. Fingerprint Gender classification

The proposed system for gender classification is built based on the fusion of fingerprint features obtained by using DWT and SVD. This section describes two different stages named as learning stage and classification stage and the KNN classifier used for the gender classification.

### 4.1 Learning Stage

The feature vector V of size 1x260 obtained by SVD and the sub band energy vector E of size 1x19 obtained by DWT are fused to form the feature vector and used in the learning stage. The fusion of feature vector V and E is done by concatenation of features that are widely used for feature level fusion. The resulting feature vector is of the size 1x279 (1x260 +1x19). The learning stage is shown in Figure 4.

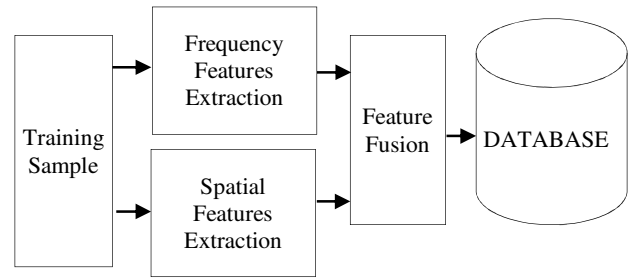

Fig. 4 Learning stage of the proposed gender classification system

The learning algorithm is as follows:
**Learning Algorithm:**
[Input] all samples of fingerprint with known class (Gender)
[Output] the feature vector of all samples as database

1) Decompose the fingerprint with 6 level decomposition of DWT.
2) Calculate the sub-band energy vector (E) using (1).
3) Calculate the Eigen vector (V) using (2).
4) Fuse the vectors E and V to form the feature vector for the particular fingerprint.
5) Insert this feature vector and the known class into the database.
6) Repeat the above steps for all the samples.

### 4.2 KNN Classifier

In pattern recognition, the k-nearest neighbour algorithm (K-NN) is the generally used method for classifying objects based on closest training examples in the feature space. K-NN is a type of instance-based learning where the function is only approximated locally and all computation is deferred until classification. In K-NN, an object is classified by a majority vote of its neighbours, with the object being assigned to the class most common amongst its k nearest neighbours (k is a positive integer, typically small). If k = 1, then the object is simply assigned to the class of its nearest neighbour. The neighbours are taken from a set of objects for which the correct classification is known. This can be thought of as the training set for the algorithm, though no explicit training step is required.

### 4.3 Classification Stage

In the classification phase, the fused feature vector of the input fingerprint is compared with the feature vectors in the database by using the KNN classifier. The distance measure used in the classifier is 'Euclidean Distance'. The classification process is as follows.

Algorithm II:  Classification Algorithm
[Input] unknown fingerprint and the feature database
[Output] the class of the fingerprint to which this unknown fingerprint is assigned
1) Decompose the given unknown fingerprint with 6 level decomposition of DWT.
2) Calculate the sub-band energy vector (E) using (2).
3) Calculate the Eigen vector (V) using (1).
4) Fuse the vectors E and V to form the feature vector for the given unknown fingerprint.
5) Apply KNN classifier and find the class of the unknown fingerprint by using the database generated in the learning phase.

## 5. Experimental Results

In this section, the performance of the proposed gender classification algorithm is verified by using the internal database. The success rate (in percentage) of gender classification using DWT, SVD and combination of both are summarized and discussed. Also, the results of the proposed method are compared with the results of earlier publications of gender classification.

### 5.1 Data set

The fingerprint images of internal database were collected by using Fingkey Hamster II scanner manufactured by Nitgen biometric solution [30], Korea. Every original image is of size 260x300 pixels with 256 grey levels and resolution of 500 dpi. The internal database includes all ten fingers collected from males and females of different ages. From the internal database, irrespective of quality and age, all ten fingers of 3570 fingerprints in which 1980 were male fingerprints and 1590 were female fingerprints are used for testing and training. These 3570 fingerprint images are separated into two sets. For the learning stage 2/3 of total images are used. The remaining images are used in the classification stage. Table 1 shows the age and gender wise samples of the internal database.

Table 1: Age and gender wise samples details

| Age Group | Male | Female | Total |
|---|---|---|---|
| Up to 12 | 70 | 60 | 130 |
| 13-19 | 190 | 320 | 510 |
| 20-25 | 1050 | 680 | 1730 |
| 26-35 | 320 | 270 | 590 |
| 36 and above | 350 | 260 | 610 |
| Total Samples | 1980 | 1590 | 3570 |

The scanned fingers were numbered as follows. Left little finger to left thumb is numbered as 1-5. Right thumb to right little finger is numbered as 6-10 as shown in Figure 5.

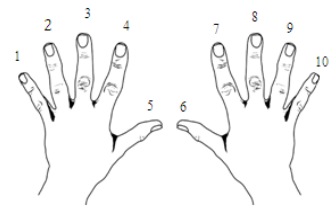

Fig. 5 Finger numbering

### 5.2 Gender classification using DWT only

The code is tested from 2nd level to 7th level and the success rate for the classification is identified. No appreciable results were obtained for the levels 2 to 4 and beyond the level 7, the results were not convincing. Significant success rate is obtained for the levels 5, 6 and 7. The subband energy vector for the level 5 is of the size 1x16 and these features are compared with templates stored in the database obtained during the learning stage. Similarly, the subband energy vectors are of the size 1x19 and 1x22 for the level 6 and 7 respectively. The results achieved by the 2-D DWT for the levels 5, 6 and 7 are listed in table 2 for each finger of the male and female.

Table 2: Gender classification rate for different levels of DWT

| Finger No. | Level 5 | | Level 6 | | Level 7 | |
|---|---|---|---|---|---|---|
| | Male | Female | Male | Female | Male | Female |
| 1 | 87.88 | 86.36 | 90.15 | 92.05 | 88.64 | 96.59 |
| 2 | 84.09 | 86.36 | 84.09 | 90.91 | 86.36 | 90.91 |
| 3 | 86.36 | 82.95 | 87.12 | 87.50 | 85.61 | 84.09 |
| 4 | 90.91 | 81.82 | 90.15 | 77.27 | 88.64 | 80.68 |
| 5 | 90.15 | 79.55 | 90.15 | 76.14 | 87.12 | 85.23 |
| 6 | 94.70 | 79.55 | 93.94 | 79.55 | 92.42 | 77.27 |
| 7 | 87.12 | 69.32 | 89.39 | 75.00 | 92.42 | 70.45 |
| 8 | 88.64 | 71.59 | 90.15 | 79.55 | 90.15 | 80.68 |
| 9 | 87.12 | 85.23 | 87.88 | 77.27 | 89.39 | 86.36 |
| 10 | 89.39 | 82.95 | 90.15 | 80.68 | 90.91 | 84.09 |
| Average | 88.64 | 80.57 | 89.32 | 81.59 | 89.17 | 83.64 |

The overall classification rate for the level 5, 6 and 7 are 84.61%, 85.46% and 86.41% respectively.

It is also observed that the success rate of right thumb finger (numbered as 6) of male are quite higher than the other fingers. Similarly, the success rate of left hand little finger (numbered as 1) of female are higher than the other fingers. The result pattern for the male and female fingers is shown in figure 6 (a) and 6(b) respectively.

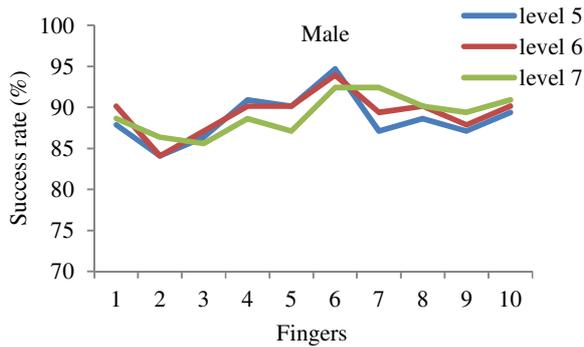

(a)

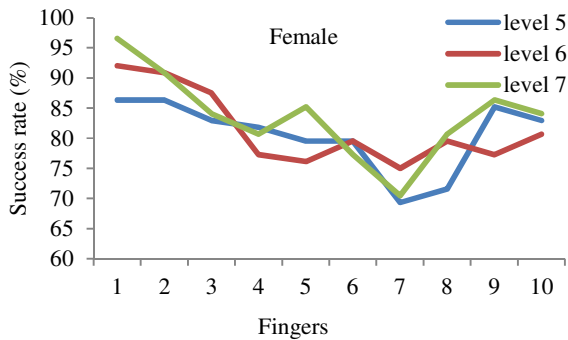

(b)

Fig 6 Result pattern of different levels of DWT (a) Male (b) Female

By average, the level 6 DWT produces the best result for male gender and the level 7 DWT produces the best result for female gender. Considering the training and testing duration, level 6 DWT is considered as the optimum level for the gender classification.

### 5.3 Gender classification using SVD only

In this section, SVD alone is applied for the gender classification and its results are compared only with the level 6 DWT. SVD generates large feature vector depends on the size of the image and for the internal database the feature vector is of the size 1x260. Thus, significant improvement on gender classification is observed and listed in table 3. The success rate for the male and female finger identification is 91.74% and 83.30% respectively and the overall classification rate is 87.52%.

Table 3: Gender classification using SVD

| Finger No. | Male | Female |
|---|---|---|
| 1 | 90.15 | 92.05 |
| 2 | 89.39 | 85.23 |
| 3 | 90.15 | 86.36 |
| 4 | 91.67 | 85.23 |
| 5 | 93.18 | 84.09 |
| 6 | 93.18 | 75.00 |
| 7 | 93.94 | 73.86 |
| 8 | 91.67 | 76.14 |
| 9 | 91.67 | 86.36 |
| 10 | 92.42 | 88.64 |
| Average | 91.74 | 83.30 |

While comparing with level 6 DWT, the SVD results are 2.64 % more for male and 2.1 % more for female gender. In SVD, thumb fingers (numbered as 5 and 6) and the right index (numbered 7) shows higher results than the other fingers. Similarly, the left little finger of the female shows higher result than the other fingers as shown in figure 7.

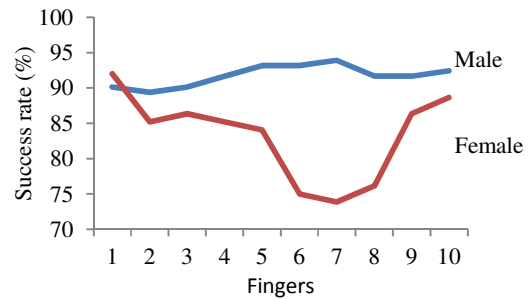

Fig. 7 Result pattern of SVD for male and female

The success rate rises from the 1st finger to 5th finger and falls from the 6th finger to 10th finger for the male fingers. But for the female fingers, the success rate falls from the 1st finger to 5th finger and rises from the 6th finger to 10th finger and thus forms like a valley structure.

### 5.4 Gender classification using combined DWT and SVD

From the table 4, it is evident that, by SVD there is an overall raise in gender classification rate of 2.35 % in comparison with the level 6 DWT. To assess the results of combined DWT and SVD, the features of these two approaches are combined by concatenation and the results are verified. To identify the better combinations of different levels of DWT and SVD, features of level 5, 6 and 7 are combined individually with SVD features and results are obtained. The gender classification results for the level 5 plus SVD, level 6 plus SVD and the level 7 + SVD are shown in the table 4. A notable increase in the success rate is achieved in each case.

Table 4: Gender classification rate of the combined SVD and DWT for the samples used for training

| Finger No. | Level 5 | | Level 6 | | Level 7 | |
|---|---|---|---|---|---|---|
| | Male | Female | Male | Female | Male | Female |
| 1 | 89.39 | 94.32 | 87.15 | 94.32 | 91.67 | 94.32 |
| 2 | 90.91 | 87.50 | 91.49 | 90.85 | 91.67 | 86.36 |
| 3 | 93.18 | 84.09 | 92.87 | 87.5 | 93.18 | 87.50 |
| 4 | 93.94 | 82.95 | 95.56 | 84.95 | 89.39 | 82.95 |
| 5 | 93.18 | 75.00 | 94.94 | 75.69 | 94.70 | 76.14 |
| 6 | 93.94 | 75.00 | 93.87 | 73.86 | 92.42 | 73.86 |
| 7 | 93.18 | 76.14 | 94.38 | 80.55 | 93.94 | 78.41 |
| 8 | 93.18 | 87.50 | 93.23 | 88.5 | 90.91 | 84.09 |
| 9 | 91.67 | 88.64 | 91.84 | 92.35 | 94.70 | 92.05 |
| 10 | 90.15 | 85.23 | 86.98 | 92.64 | 90.15 | 88.64 |
| Average | 92.27 | 83.64 | 92.23 | 86.12 | 92.27 | 84.43 |

**Selection of optimum DWT level:** The overall classification rate for the combined SVD and DWT level 5, 6 and 7 are 87.96%, 89.16% and 88.35% respectively for the samples used for the training. While testing this method for external input other than the samples used for training, the overall classification rate obtained is 86.06%, 88.28% and 86.16% for SVD and DWT level 5, 6 and 7 respectively. Thus, level 6 gives greater success than level 5 and level 7. The results are shown in table 5.Thus the level 6 DWT is considered as an optimum level for the gender identification.

Table 5: Gender classification rate of the combined SVD and DWT for the testing samples

| Finger No. | Level 5 | | Level 6 | | Level 7 | |
|---|---|---|---|---|---|---|
| | Male | Female | Male | Female | Male | Female |
| 1 | 86.99 | 91.67 | 87.67 | 94.32 | 87.67 | 90.63 |
| 2 | 89.04 | 86.46 | 89.86 | 90.88 | 89.04 | 83.33 |
| 3 | 90.41 | 81.25 | 92.79 | 85.88 | 91.10 | 86.46 |
| 4 | 93.15 | 79.17 | 95.46 | 79.68 | 89.73 | 78.13 |
| 5 | 91.78 | 73.96 | 94.92 | 75.92 | 93.15 | 72.92 |
| 6 | 93.84 | 72.92 | 92.8 | 73.87 | 93.15 | 71.88 |
| 7 | 89.73 | 77.08 | 93.17 | 80.69 | 91.78 | 78.13 |
| 8 | 91.78 | 83.33 | 92.57 | 84.28 | 91.10 | 79.17 |
| 9 | 89.73 | 86.46 | 90.73 | 89.78 | 93.15 | 87.50 |
| 10 | 86.99 | 85.42 | 86.76 | 93.58 | 87.67 | 87.50 |
| Average | 90.34 | 81.77 | 91.67 | 84.89 | 90.75 | 81.56 |

Gender classification rate of all the approaches discussed above are shown as a bar chart in figure 8 and 9.

Figure 8 shows the increased rate of gender identification by the combined DWT and SVD for male than the individual approach of DWT and SVD. The success rate is more for the thumb and index fingers than other fingers.

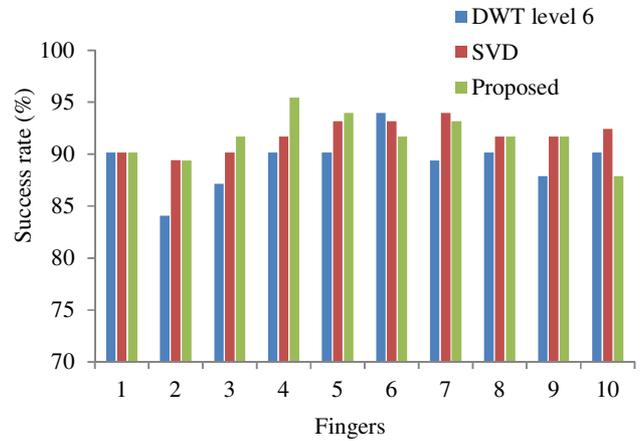

Fig. 8 Male gender classification rate of DWT, SVD and combined DWT and SVD

Figure 9 shows the increased rate of gender identification by the combined DWT and SVD for female than the individual approach of DWT and SVD. The success rate is more for the left index (numbered 1) than other fingers.

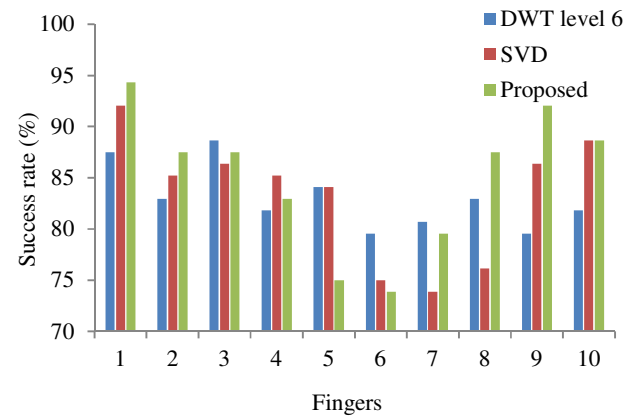

Fig. 9 Female gender classification rate of DWT, SVD and combined DWT and SVD

Average success rate for the methods discussed are shown as line chart in the figure 10. There is raise in success rate in proposed method than DWT alone by 2.56% and only 0.08% less with SVD alone for male. But for female, the proposed method produces 2.29% and 1.87% more than DWT and SVD alone. Similar to the DWT and SVD, the left hand thumb (numbered as 5) and the left hand index finger (numbered as 4) shows higher success rate for male and the left hand little finger (numbered as 1) for female shows higher success rate than the other fingers.

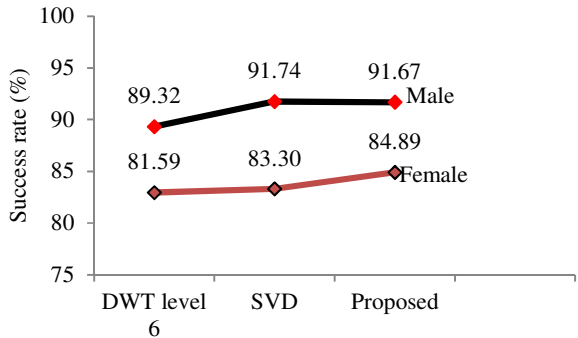

Fig. 10 Performance comparison of proposed method

The success rate rises from the 1st finger to 5th finger and falls from the 6th finger to 10th finger for the male fingers and thus the pattern is projected in the middle area. But for the female fingers, the success rate falls from the 1st finger to 5th finger and rises from the 6th finger to 10th finger and thus the result pattern forms like a valley in the middle. The result pattern of table 4 and 5 are shown in figure 11 and 12 respectively.

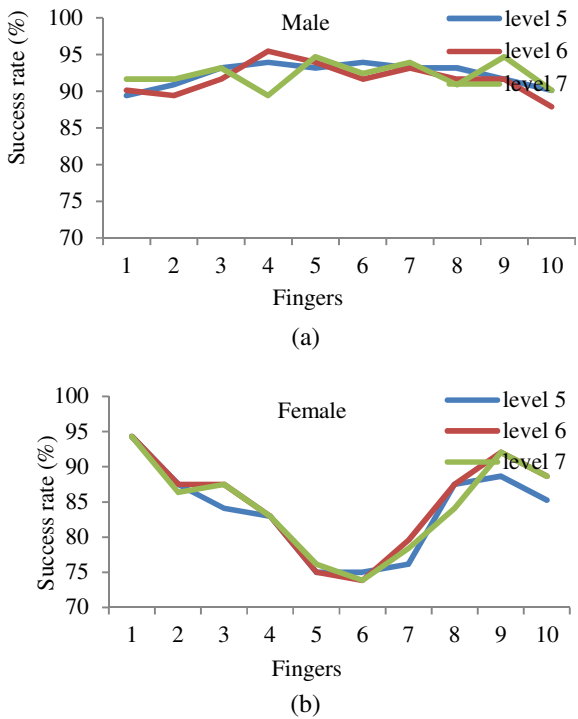

Fig. 11 Result patterns of table 4, (a) Male (b) Female

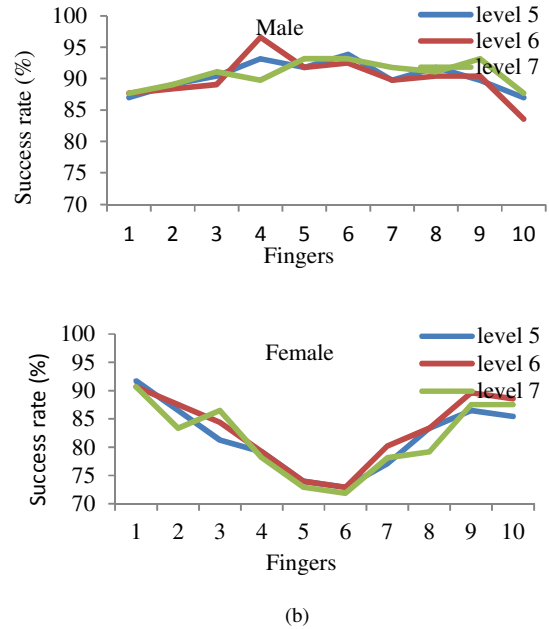

Fig. 12 Result patterns of table 4, (b) Result patterns of table 5.

## 5.6 Performance comparison

In this sub-section, the proposed method is compared with the earliest published results. The results of Ahmed Badawi et al. [10], is compared and in their study, ridge thickness to valley thickness ration (RTVTR), ridge count, white lines count, ridge count asymmetry and pattern type concordance were used as features and FCM, LDA, and NN classifiers were used for gender classification. For this study The RTVTR, and white lines count features were analyzed for 255 persons (150 males, and 105 females). These images were scanned from the person's ink print. Manish Verma et al. [11], in their paper used ridge density and ridge width in addition to RTVTR as features and results were obtained for the internal database of 200 male and 200 female fingerprints with SVM classifier. Gender classification accuracies of the proposed method and the published results are shown in table 6.

Table 6: Comparison of gender classification accuracies

| | Ahmaed Badawi et al. | | | Manish Verma et al. | Proposed method |
|---|---|---|---|---|---|
| Features used | RTVTR, white line count, ridge count asymmetry pattern type | | | RTVTR, Ridge width and ridge density | DWT & SVD |
| Classifiers | FCM | LDA | NN | SVM | KNN |
| Male | 58.67 | 96.15 | 90.38 | 86 | 91.67 |
| Female | 56.33 | 72.97 | 83.78 | 90 | 84.89 |
| | 56.47 | 84.52 | 87.64 | 88 | 88.28 |

## 6. Conclusions

In this work, we have proposed a new method for gender classification of fingerprint images based on level 6 DWT and SVD. This method considered the frequency features of the wavelet domain and the spatial features of the singular value decomposition. The spatial features include the internal structure of the fingerprint images and the fusion of these features with the frequency features produces improved performance in gender classification. The level 6 DWT is selected as optimum level for the gender classification by analysing the results obtained for the database used for training and testing and the database used other than the training and testing. By the proposed method, the gender classification rate achieved is 91.67% for male and 84.89% for female.

For the finger-wise gender classification, the success rate is higher for the little fingers and decreases from little fingers to thumb fingers. The success rates falls at the rate of 2.56% minimum to 8.05% maximum from the finger 1 to 5 and rises at the rate of 1.32% to 8% from finger 6 to 10. Thus the result pattern shown in line diagrams formed like a valley. Similarly among the male fingers the success rate is higher for the thumb fingers and index fingers and decreases from the thumb to little fingers. The success rates rises at the rate of 0.77% minimum to 7.8% maximum from the finger 1 to 5 and falls at the rate of 0.75% minimum to 4.38% maximum from finger 6 to 10. Thus the result patterns shown in line diagrams are slightly projected at the middle.

Our future work is to extend the proposed method of gender classification using the spatial parameters. Also, it is aimed to use various other techniques to increase the success rate.

**P. Gnanasivam** received his A.M.I.E in electronics and communication engineering from Institution of Engineers (India) in 1989 and M.E in Applied Electronics from the Anna University, India in 2003. He is currently working towards the Ph.D. degree in the Department of Electronics and Communication Engineering, Anna University, India. His research interests include Pattern recognition, Image processing, Biometrics and Embedded system.

**Dr. S. Muttan**, Professor, Centre for Medical Electronics, Department of ECE, Anna University, former Asst. secretary of Indian Association of Biomedical Scientists and also a Life member of Biomedical society of India, Life Member in ISTE. He has been guiding UG and PG students and research scholars on various fields in Electronics communications and Information and Communication Technology applied to medicine. His area of research includes Medical Informatics, pattern recognition, and biometrics and e-health services. He has published many research papers in both National and International conferences and Journals. He completed his PhD. in Evolution and Design of Integrated Cardiac Information system in Multimedia.